\documentclass[10pt,twocolumn]{ICCAS2021}
 
\usepackage{diagbox}
\usepackage{cite}
\usepackage{amsmath,amssymb,amsfonts}
\usepackage{algorithmic}
\usepackage{graphicx}
\usepackage{textcomp}

\begin{document}

\title{Indoor Path Planning for an Unmanned Aerial Vehicle via Curriculum Learning}

\author{Jongmin Park${}^{1}$, Sooyoung Jang${}^{2}$, and Younghoon Shin${}^{1*}$ }

\affils{ ${}^{1}$School of Integrated Technology, Yonsei University, \\
Incheon, 21983, Korea (jm97, yh.s@yonsei.ac.kr) \\
${}^{2}$Electronics and Telecommunications Research Institute, \\
Daejeon, 34129, Korea (sy.jang@etri.re.kr) \\
{\small${}^{*}$ Corresponding author}}

%\thanks{ \noindent
%   This paper is supported by my funding agencies.
%  }

\abstract{
In this study, reinforcement learning was applied to learning two-dimensional path planning including obstacle avoidance by unmanned aerial vehicle (UAV) in an indoor environment. The task assigned to the UAV was to reach the goal position in the shortest amount of time without colliding with any obstacles. Reinforcement learning was performed in a virtual environment created using Gazebo, a virtual environment simulator, to reduce the learning time and cost. Curriculum learning, which consists of two stages was performed for more efficient learning. As a result of learning with two reward models, the maximum goal rates achieved were 71.2\% and 88.0\%.
}


\maketitle

%-----------------------------------------------------------------------

\section{Introduction}
Unmanned aerial vehicles (UAVs) are being studied and used in various fields \cite{Alladi20:SecAuthUAV, Lee20191187, Singh20:On, Lee2018:Simulation}. In the case of a quadcopter \cite{Xuan-Mung20, Shin2017617, Talaeizadeh21, Kim20141431}, which is one of the most common types of UAV, its position can be maintained through hovering, which is not possible with a fixed-wing UAV. Various sensors can be mounted on the UAV and the location of the UAV can be determined using global navigation satellite systems (GNSS) \cite{Causa21, Sun2020889, Park2021919, Savas21, Park2018387}, long-term evolution (LTE) based positioning \cite{Shamaei21, Maaref20, Kang20191182, Jeong2020958, Lee2020:Preliminary, Lee2020939, Lee20202347, Jia21:Ground, Kang2020774}, enhanced long-range navigation (eLoran) \cite{Son20191828, Son2018666, Williams13, Kim2020796, Qiu10, Park2020824, Hwang2018, Li20, Rhee21:Enhanced}, and other techniques \cite{Kim2017:SFOL, Rhee2019, Park2020800}. UAVs can be used for target searching, weather information acquisition, aerial photography, delivery, communication repeating, and for entertainment using light sources \cite{Sibanyoni19, Dorling17, Hiraguri20}.

Considering the possibilities of using such UAVs, research is being conducted to optimize the movement of UAVs using artificial intelligence (AI) \cite{Kim2020784, Kouroshnezhad21, Chhikara21}. Specifically, reinforcement learning, which has been widely investigated with the recent development in deep learning, was used in \cite{Kim2020784}. Reinforcement learning involves learning the optimal behavior in a given situation through actions and rewards and is mainly used in robots and game AI.

In this study, we performed reinforcement learning to ensure that a UAV could reach the goal position in the shortest amount of time while avoiding obstacles. In consideration of learning time and cost, the learning was conducted in a virtual indoor environment created using Gazebo \cite{Koenig04}, a virtual environment simulator. In addition, for the two reward models that we proposed, curriculum learning was performed to increase the efficiency of learning. First, learning was conducted in an environment without obstacles, then after learning had progressed to a certain level, obstacles were added, and the learning was continued.

Curriculum learning is a machine learning technique that involves learning simple tasks sufficiently and then progressing to difficult and complex tasks. This technique offers the advantages of generalization and a fast convergence speed \cite{Bengio09}. In our study, learning was first performed for a simple path planning to ensure that a UAV could fly quickly to a goal point in an environment without obstacles. After this simple task was learned, the learning was performed in an environment with obstacles to train the UAV to fly to a goal point within a short time while avoiding obstacles. Such curriculum learning is more efficient than learning a difficult task from the beginning.

\section{Virtual environment}
\begin{figure}
  \centering
  \includegraphics[width=1\linewidth]{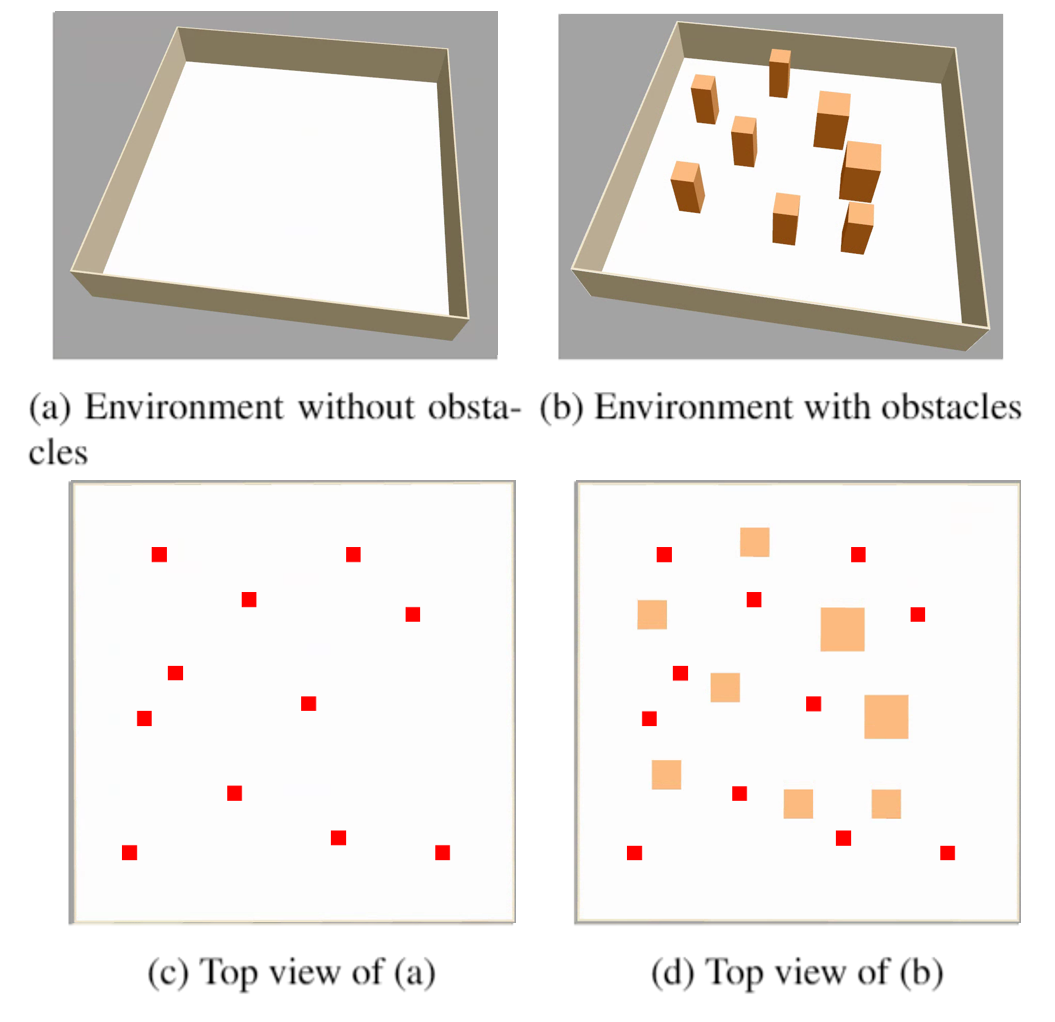}
  \caption{Virtual indoor environment}
  \label{fig:indEnv}
\end{figure}

To implement the UAV virtual environment and learning environment, Gazebo, Robot Operating System (ROS) \cite{Quigley09}, and OpenAI Gym \cite{Brockman16} were used. Using the building editor in Gazebo, a virtual indoor environment of $30\, \rm{m} \times 30\, \rm{m}$ with a few obstacles was created. Fig. \ref{fig:indEnv}(a) shows the virtual indoor environment without obstacles used at the beginning of learning, and Fig. \ref{fig:indEnv}(b) shows the a virtual indoor environment with obstacles. Figs. \ref{fig:indEnv}(c) and \ref{fig:indEnv}(d) present top view images of the environment in Figs. \ref{fig:indEnv}(a) and \ref{fig:indEnv}(b), respectively. The red squares indicate the coordinate set of the UAV. The UAV randomly selects two coordinates from this set as the starting point and goal point for training.

ROS is a meta-operating system for robots that includes hardware abstraction, low-level device control, implementation of commonly used functionality, message passing between processes, and package management \cite{Quigley09}. We used ROS because it can be easily integrated into other robot software frameworks, and many studies on robots or UAVs have utilized ROS. The action of the UAV in the OpenAI Gym learning environment can be transferred to the virtual environment implemented in Gazebo; moreover, information from the sensor mounted on the virtual UAV, such as whether the UAV has collided, can be delivered to the Gazebo learning environment through message passing. 

OpenAI Gym is a toolkit for reinforcement learning research \cite{Brockman16}. Various learning environments have already been implemented, and new learning environments can be created as per the OpenAI Gym format. In addition, OpenAI Gym is convenient to link with TensorFlow or RLlib \cite{Liang18}. In this study, a learning environment was created based on the OpenAI Gym format, and learning was performed by linking OpenAI Gym with RLlib.

\section{Learning environment}
The time unit for selecting an action in the current state and receiving a reward is called a \textit{step}, and it is the smallest unit of learning. An \textit{episode} consisting of several steps refers to the time from when the UAV starts the task until it reaches the goal point or collides with an obstacle. When the learning progresses beyond a certain number of episodes, it becomes one \textit{iteration}, and the model parameters of reinforcement learning are updated every iteration.

A \textit{train batch size} of RLlib, which determines the size of one iteration, was set to 10,000 steps in this study, and the learning was started in the environment without obstacles for 200 iterations. The learning was continued in the environment with obstacles for 100 iterations.
 
The reinforcement learning algorithm used in this study is a proximal policy optimization (PPO) algorithm \cite{Schulman17}, and is provided with PPOTrainer in RLlib. PPO is a policy gradient-based reinforcement learning method that is more suitable for problems with a continuous state space than for Q-learning-based reinforcement learning such as deep Q-network (DQN) \cite{Volodymyr13, Jang19, Klissarov17}. Because the state space in this study was continuous, PPO was selected as the learning algorithm. 
In this study, learning rate, named \textit{lr} in RLlib, was set to $5 \times 10^{-5}$, trace-decay parameter, named \textit{lambda} in RLlib, was set to 1, and initial coefficient for Kullback-Leibler (KL) divergence, named \textit{kl\_coeff} in RLlib, was set to 0.2.

Reinforcement learning is the process of studying the action that maximizes the reward in the current state. Thus, the performance of learning is determined by the state space, action space, and reward model.

\subsection{State space}
The state space in our study is divided into three types: heading, distance, and lidar data. Heading in this paper refers to the difference in angle between the straight line connecting the UAV and the goal and the heading direction of the current UAV (in radian). Distance refers to the 2D Euclidean distance between the UAV and the goal. Lidar data represents information obtained from a lidar mounted on a UAV.

\subsection{Action space}
The action space was divided into three forward linear velocities and five yaw rates, and a total of 15 actions were set. The three forward linear velocities were 1 m/s, 0.5 m/s, 0 m/s, and the five yaw rates were $-2/12$ rad/s, $-1/12$ rad/s, 0 rad/s, 1/12 rad/s, and 2/12 rad/s. A negative yaw rate indicates turning counterclockwise, while a positive yaw rate indicates turning clockwise. Since the UAV moved in a 2D space, its vertical velocity was set to zero.

\subsection{Reward model}
Because the reward model is the factor that can have the greatest impact on learning performance, two reward models were designed, and the learning performance was compared between these models. Our reward model is divided into terminal reward, time penalty, progress distance, and progress heading. A difference exists between the two reward models in terms of the progress heading.

Terminal reward is the reward given at the end of the episode. If the task is successful, a reward of $+2000$ is given, and if the task is failed, a reward of $-500$ is given. The time penalty is for performing a task within the shortest amount of time, and a reward of $-1$ is given to each step. Progress distance is a value obtained by multiplying 40 by the difference in the Euclidean distance between the UAV and the goal in the previous step and the current step; it has a positive value when the UAV approaches the goal and a negative value when it moves away from the goal. Progress heading is a reward that varies depending on the heading in the state space; when the absolute value of the heading is less than 20 degrees, a value obtained by multiplying the linear speed by 5 is given as a reward for moving quickly to the goal. In addition, when the absolute value of the heading is greater than 20 degrees, the reward is given by multiplying $\frac{45}{17} (\frac{|\mathrm{heading}|}{\pi} - \frac{1}{18})$, which is a linear function of heading, by $-(1 +  \mathrm{linear \: speed})$ for reward model 1 and $-(1 + 3 \times \mathrm{linear \: speed})$ for reward model 2 to reduce the forward linear velocity and ensure that the UAV heads toward the goal.

\section{Simulation results}

\subsection{Environment without obstacles}
\begin{figure}
  \centering
  \includegraphics[width=1.0\linewidth]{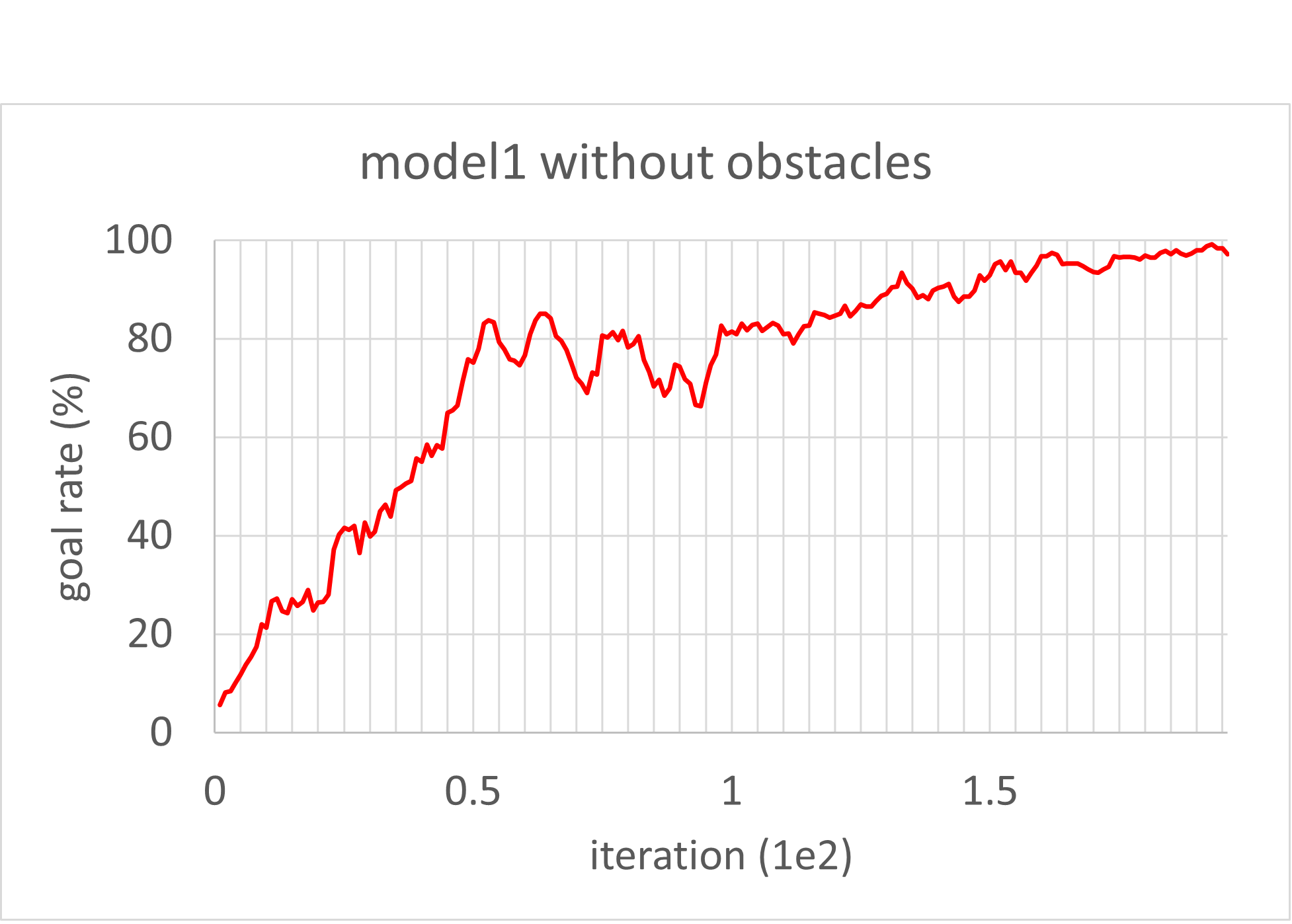}
  \caption{Moving average of goal rate for reward model 1 in the environment without obstacles}
  \label{fig:Model1woObs}
\end{figure}

\begin{figure}
  \centering
  \includegraphics[width=1\linewidth]{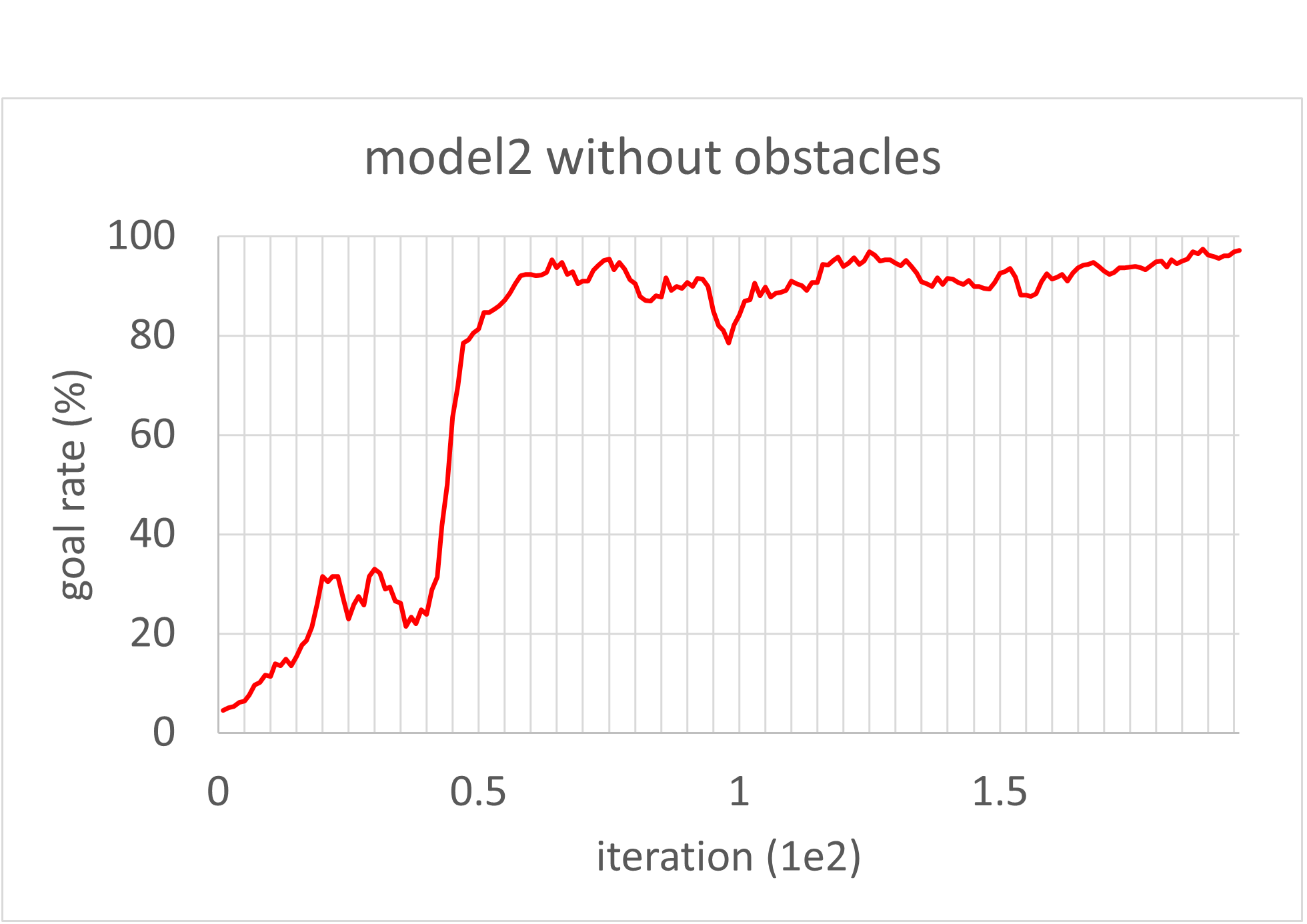}
  \caption{Moving average of goal rate for reward model 2 in the environment without obstacles}
  \label{fig:Model2woObs}
\end{figure}

Figs. \ref{fig:Model1woObs} and \ref{fig:Model2woObs} show the moving average of goal rate for reward models 1 and 2, respectively, with learning trained for 200 iterations in an environment without obstacles. The moving average was calculated based on the goal rates of the recent five iterations. 
Reward models 1 and 2 achieved a goal rate of 95.8\% and 94.4\%, respectively, in the 200 iterations.

\subsection{Environment with obstacles}

\begin{figure}
  \centering
  \includegraphics[width=0.9\linewidth]{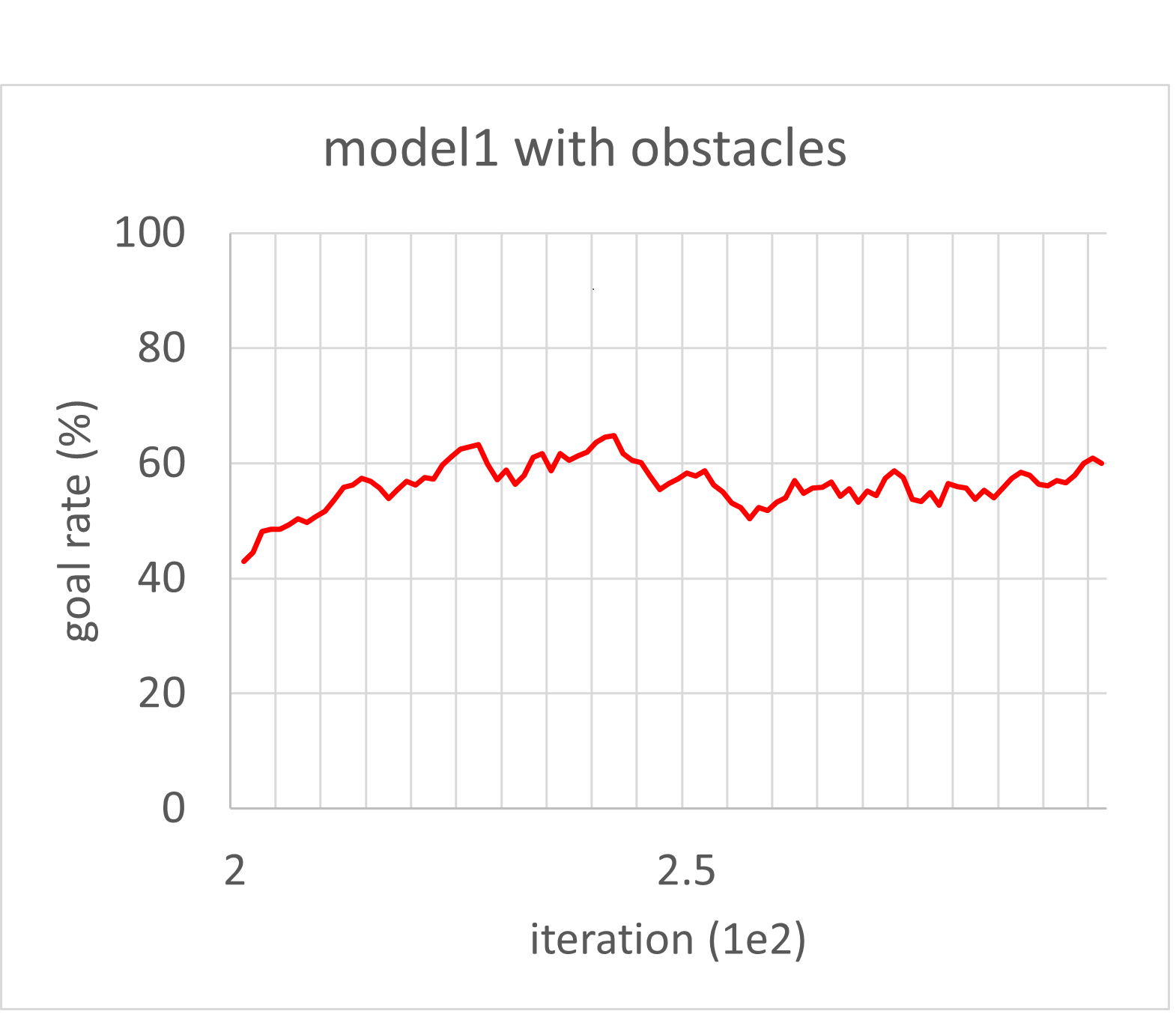}
  \caption{Moving average of goal rate for reward model 1 in the environment with obstacles}
  \label{fig:Model1wObs}
\end{figure}

\begin{figure}
  \centering
  \includegraphics[width=0.9\linewidth]{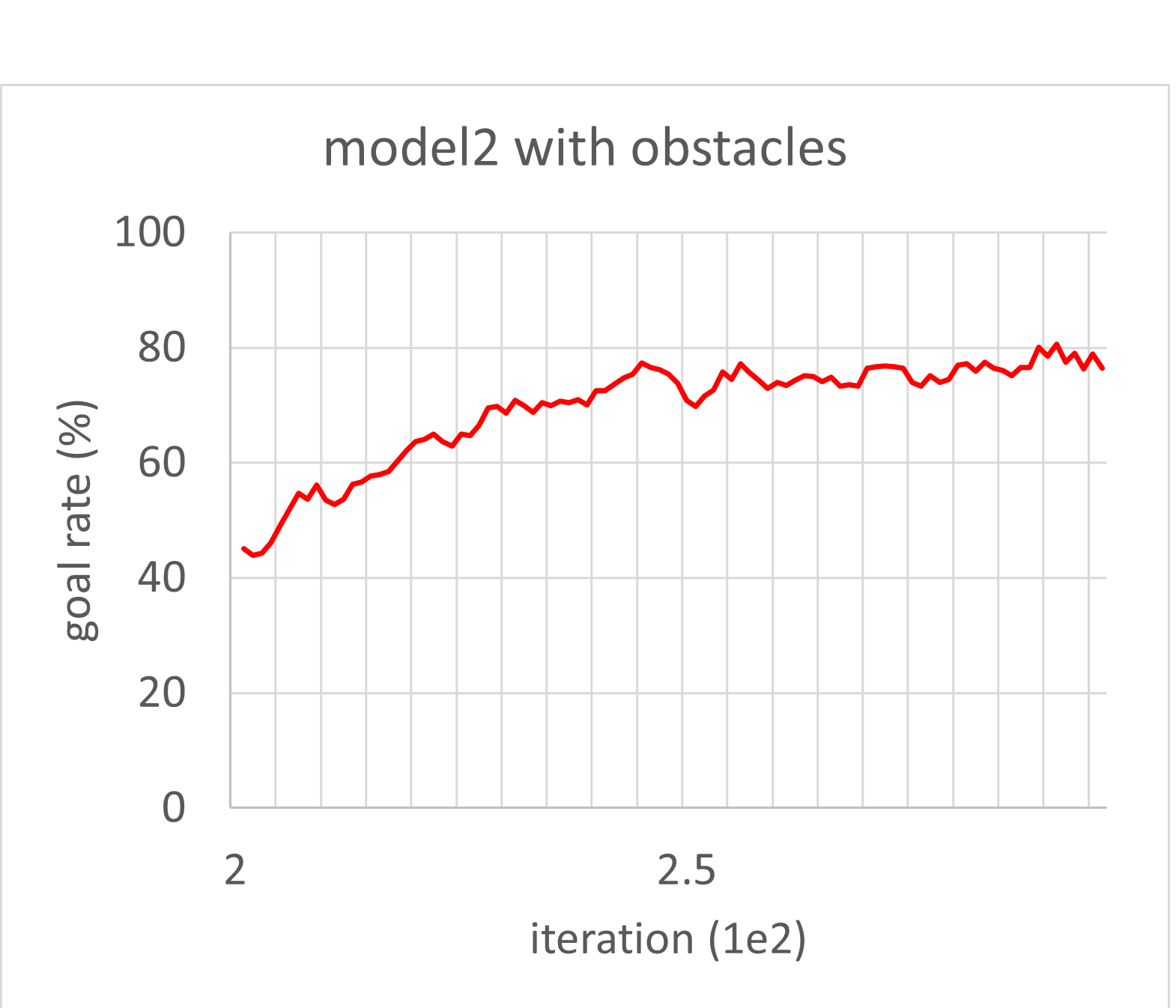}
  \caption{Moving average of goal rate for reward model 2 in the environment with obstacles}
  \label{fig:Model2wObs}
\end{figure}

Figs. \ref{fig:Model1wObs} and \ref{fig:Model2wObs} show the moving average of goal rate for reward models 1 and 2, respectively, which additionally learned for 100 iterations in an environment with obstacles. Reward models 1 and 2 achieved a maximum goal rate of 71.2\% and  88.0\%, respectively. 

\section{Conclusion}
In this study, we investigated the path planning of a UAV via reinforcement learning, including curriculum learning for two reward models. After learning for 200 iterations in an environment without obstacles, both reward models achieved a high goal rate of approximately 95\%. To proceed with curriculum learning, obstacles were added to the environment, and the learning was continued. In the environment with obstacles, the goal rate dropped to approximately 30--40\% and then gradually increased again. For reward models 1 and 2, the maximum goal rates were 71.2\% and 88\%, respectively; thus, reward model 2 outperformed reward model 1. Accordingly, the UAV that learned using reward model 2 reached the goal relatively quickly, without being significantly affected by its initial heading.

\section*{ACKNOWLEDGEMENT}
This work was supported by Electronics and Telecommunications Research Institute (ETRI) grant funded by the Korean government [21ZR1100, A Study of Hyper-Connected Thinking Internet Technology by Autonomous Connecting, Controlling and Evolving Ways].

\bibliographystyle{IEEEtran}
\bibliography{mybibfile, IUS_publications}

\end{document}